\documentclass[lettersize,journal]{IEEEtran}
\usepackage{amsmath,amsfonts}
\usepackage{algorithmic}
\usepackage{algorithm}
\usepackage{array}
\usepackage{textcomp}
\usepackage{stfloats}
\usepackage{url}
\usepackage{verbatim}
\usepackage{graphicx}
\usepackage{booktabs}
\usepackage{wrapfig}
\usepackage{caption} 
\usepackage{lipsum}  
\usepackage{hyperref}
\usepackage{url}
\usepackage{amsmath,amssymb,amsthm,enumitem,longtable,graphicx,bm,subcaption,float, color, bbm, booktabs,multirow}
\usepackage{titletoc}
\usepackage{cite}

\usepackage{tabularx}
\usepackage{threeparttable} 

\newif\ifanonsubmission
\anonsubmissionfalse 

\begin{document}

\title{Learning Representations from Heterogeneous Data for Robust Heart Rate Modeling}

\ifanonsubmission
  \author{Anonymous Authors}
\else
\author{%
\IEEEauthorblockN{%
Zhengdong Huang\IEEEauthorrefmark{1}\footnotemark[1]
\thanks{[1] Equal contribution. Names are listed in alphabetical order.}, %
Zicheng Xie\IEEEauthorrefmark{1}\footnotemark[1], %
Wentao Tian\IEEEauthorrefmark{1}, %
Jingyu Liu\IEEEauthorrefmark{2}, %
Lunhong Dong\IEEEauthorrefmark{3}\footnotemark[2], %
Peng Yang\IEEEauthorrefmark{1}\footnotemark[2]
\thanks{[2] Corresponding author}
\IEEEauthorblockA{\IEEEauthorrefmark{1}\IEEEauthorrefmark{2}\IEEEauthorrefmark{3}Southern University of Science and Technology, Shenzhen, China}
}
}
\fi



\maketitle

\begin{abstract}
Heart rate prediction is vital for personalized health monitoring and fitness, while it frequently faces a critical challenge in real-world deployment: \emph{data heterogeneity}. We classify it in two key dimensions: \emph{source heterogeneity} from fragmented device markets with varying feature sets, and \emph{user heterogeneity} reflecting distinct physiological patterns across individuals and activities. Existing methods either discard device-specific information, or fail to model user-specific differences, limiting their real-world performance. To address this, we propose a framework that learns latent representations agnostic to both heterogeneity, enabling downstream predictors to work consistently under heterogeneous data patterns. Specifically, we introduce a random feature dropout strategy to handle source heterogeneity, making the model robust to various feature sets. To manage user heterogeneity, we employ a history-aware attention module to capture long-term physiological traits and use a contrastive learning objective to build a discriminative representation space. To reflect the heterogeneous nature of real-world data, we created a new benchmark dataset, \textsc{ParroTao}. Evaluations on both \textsc{ParroTao} and the public FitRec dataset show that our model significantly outperforms existing baselines by 17.5\% and 10.4\% in terms of test MSE, respectively. Furthermore, analysis of the learned representations demonstrates their strong discriminative power, and two downstream application tasks confirm the practical value of our model.

\end{abstract}

\def\abstractname{Note to Practitioners}
\begin{abstract}
In fitness monitoring, wearable devices record different sensor signals across manufacturers, and individuals show distinct physiological responses to exercise. These heterogeneities challenge building reliable heart rate prediction systems.  Our work addresses this challenge by learning a unified data representation that is robust to inconsistent sensor channels and adaptive to individual physiological profiles, employing random feature dropout for device-agnostic modeling, history-aware attention for capturing long-term fitness traits, and contrastive learning for distinguishing users and activities. Evaluated on a public benchmark and our multi-device \textsc{ParroTao} dataset, the model significantly outperforms baselines. Beyond forecasting, it enables personalized route recommendation, helping athletes for training routes selection and heart rate imputation. The framework facilitates personalized training plan generation, real-time intensity monitoring, and cross-platform deployment in heterogeneous wearable ecosystems.

\end{abstract}

\begin{IEEEkeywords}
Heart Rate Prediction, Rrepresentation Learning, Contrastive Learning.
\end{IEEEkeywords}

\section{Introduction}
\label{sec:intro}

\IEEEPARstart{H}{eart} rate is a key marker of cardiorespiratory health and physical performance. Accurate heart rate prediction helps individuals and clinicians monitor workload, detect risk, and plan activities with greater confidence. For fitness enthusiasts, it links training plans to physiological response by estimating how the heart will respond to a chosen pace or interval structure, which helps users target appropriate zones, adjust intensity in real time, and avoid overreaching~\cite{modeling2023personalized,sumida2013estimating}. For professional athletes, predicted heart rate profiles support evaluation of training effectiveness across sessions and inform recovery management, reducing the chance of injury~\cite{Zadeh2021Predicting,arnold2017wearable}. For older adults and patients in rehabilitation, heart rate prediction can flag abnormal fluctuations in advance and guide timely intervention, which improves safety during daily activities and exercise~\cite{ballinger2018deepheart,dunn2021wearable, 7270350}.

Early heart rate prediction models were primarily physiology-based, employing coupled ordinary differential equations to represent cardiovascular dynamics~\cite{stirling2008model,zakynthinaki2015modelling,mazzoleni2016modeling,mazzoleni2018dynamical,engelen1996effects}. While these models offer interpretability and perform well in controlled settings, they are typically developed on small, homogeneous cohorts under strict laboratory protocols. This reliance on controlled conditions limits their robustness and generalization to free-living populations~\cite{modeling2023personalized}. The recent proliferation of wearable devices has enabled the collection of large-scale, real-world data, paving the way for data-driven approaches. Consequently, researchers have increasingly applied machine learning methods, from statistical models~\cite{oyeleye2022predictive,fang2021bayesian} to advanced architectures like recurrent and attention-based neural networks~\cite{arima2022timeseries,alharbi2021real,lin2023new, 11016077}. These data-driven techniques capture complex temporal patterns, model nonlinear relationships, and integrate multiple modalities, leading to more accurate and robust predictions in real-world conditions.

\begin{figure}[!t]
\centering
\begin{subfigure}[t]{\linewidth}
    \centering
    \includegraphics[width=\linewidth]{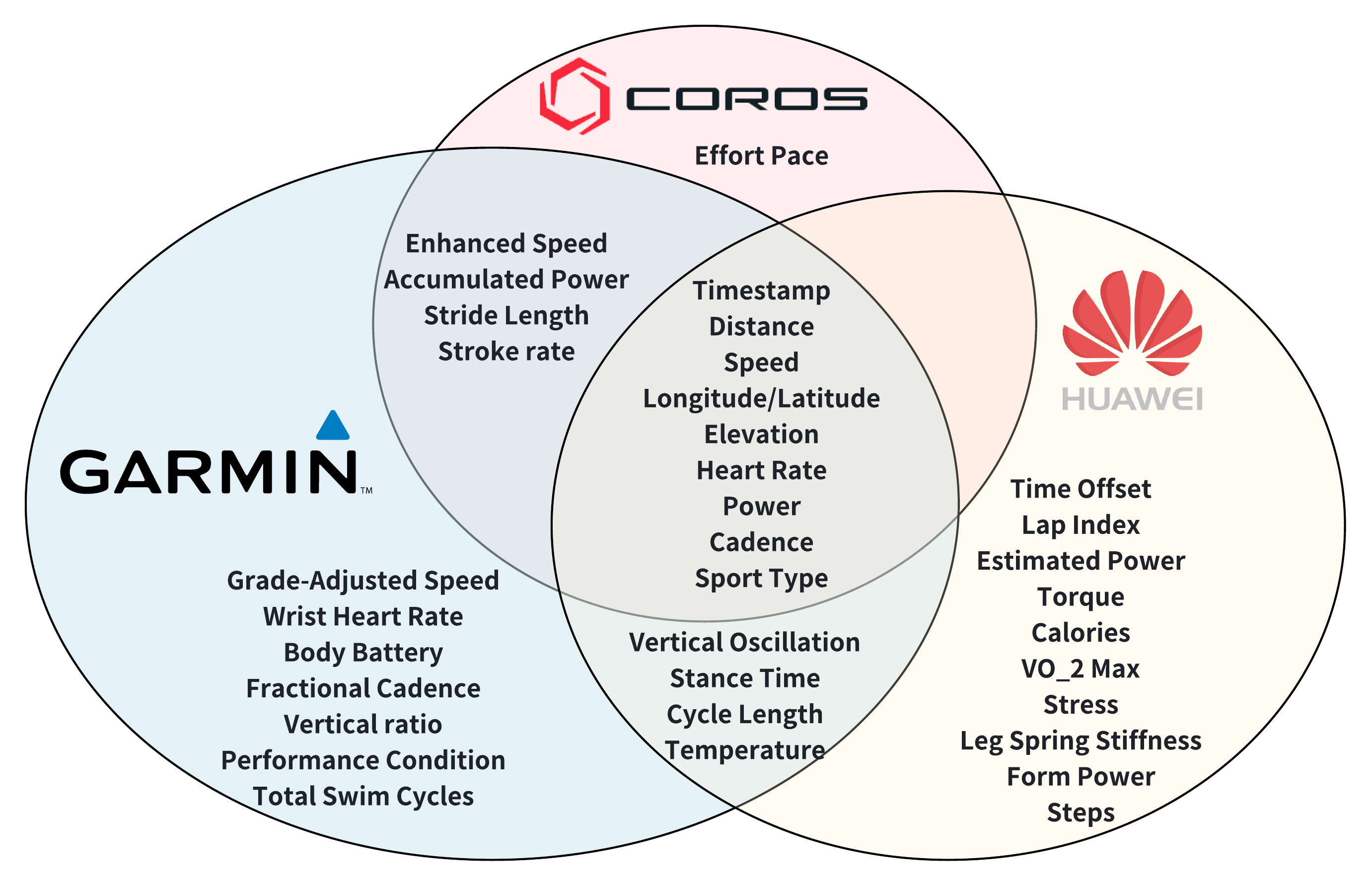}
    \caption{Device-level differences in feature sets.}
    \label{fig:heterogeneity_a}
\end{subfigure}
\hfill
\begin{subfigure}[t]{\linewidth}
    \centering
    \includegraphics[width=\linewidth]{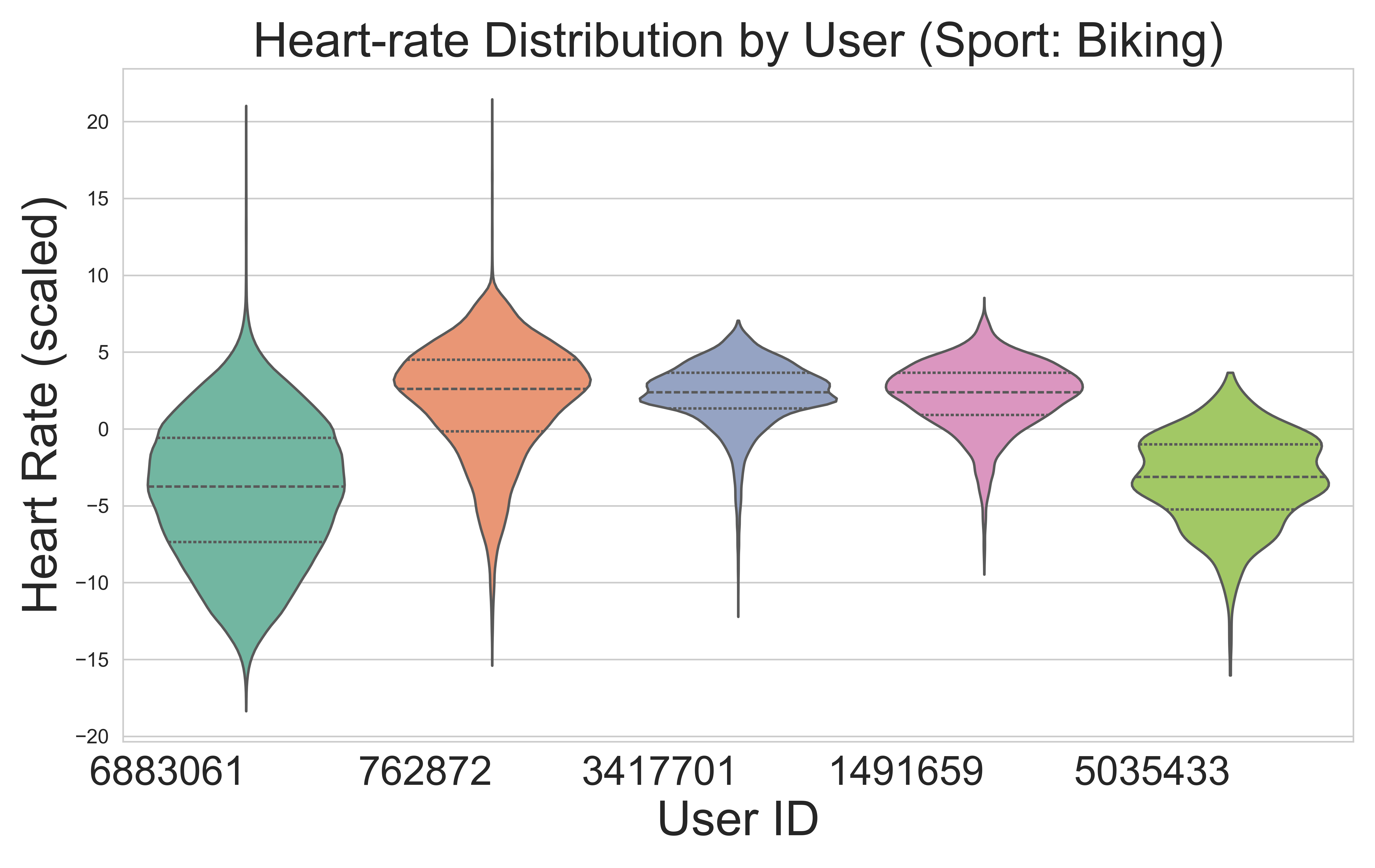}
    \caption{Heart rate distributions during biking for five users from FitRec dataset~\cite{ni2019modeling}.}
    \label{fig:heterogeneity_b}
\end{subfigure}
\caption{Data heterogeneity in wearable data. (a) Three popular wearable devices—Garmin Forerunner 255, Coros Pace 2, and Huawei GT 2—capture different feature sets, indicating source heterogeneity. (b) Users show distinct heart rate distributions under the same activity, highlighting user heterogeneity.}
\label{fig:heterogeneity}
\end{figure}

Despite this progress, \emph{data heterogeneity} remains a significant challenge for real-world deployment~\cite{9944935, 10128148}. This challenge stems from the necessity for systems to process data from diverse devices and users, leading to two primary forms of heterogeneity: \emph{source heterogeneity} and \emph{user heterogeneity}. \emph{Source heterogeneity} arises from a fragmented device market. Data from different devices or datasets often provide different feature sets and are sampled at different temporal resolutions, even for the same activity (Fig.~\ref{fig:heterogeneity_a}). \emph{User heterogeneity} reflects physiological and behavioral diversity. Even with the same device, sensed features can vary across activity types, and individuals exhibit distinct physiological profiles that lead to different heart rate distributions for the same activity~\cite{10740338} (Fig.~\ref{fig:heterogeneity_b}). Most existing methods treat such heterogeneous data as if they were homogeneous. Many are trained and evaluated on a single dataset, which limits their applicability when integrating multiple sources. When datasets are combined, a common practice is to keep only the intersection of available features and to unify the temporal resolution to the coarsest sampling rate~\cite{11016064}. This discards device-specific signals and can degrade predictive accuracy. In addition, current methods seldom leverage full historical information to model differences across users and activities, which limits personalization.

To address these challenges of data heterogeneity, we propose to learn a \emph{unified representation space} for heterogeneous heart rate time series, where a robust predictive model can be trained. For this purpose, we need to project disparate data points—originating from heterogeneous devices, users, and activities—into a common latent space. To address \emph{source heterogeneity}, we employ a flexible encoder with a random feature dropout strategy. This technique enables the model to map data with potentially different feature sets into a shared representation space. 
To tackle \emph{user heterogeneity}, we introduce two components to enhance personalization and discriminative power. First, a history-aware attention module captures user-specific physiological traits by selectively weighing historical data. Second, we design a contrastive loss to learn discriminative representations. This forces the representations of different users and activities to be distinct while ensuring that representations for the same user and activity remain similar. By combining these elements, our approach generates a powerful, personalized, and robust embedding for more accurate heart rate prediction.

Apart from the proposed method, although data heterogeneity is common in real-world scenarios, most existing public benchmarks rarely account for cross-source and cross-user variations. To bridge this gap, we have constructed and publicly released \textsc{ParroTao}—a large-scale, multi-device, multi-activity dataset that reflects the heterogeneous nature of real-world data. This dataset comprises recorded exercise segments from recreational athletes, captured by devices from diverse manufacturers during a wide range of athletic activities. Unlike common practices that unify datasets by retaining only the intersection of features, we deliberately preserve the distinct, device-specific feature sets. This design, combined with the variety of activities, ensures our dataset more faithfully represents the challenges of real-world data, making it a more rigorous benchmark for evaluating a model's performance. On both the public FitRec dataset~\cite{ni2019modeling} and our new \textsc{ParroTao} dataset, our experiments demonstrate that the proposed approach significantly outperforms existing baselines by 17.5\% and 10.4\% in terms of MSE, respectively, while also achieving superior performance across fine-grained sport categories. Furthermore, analyses of the learned representations confirm that our framework effectively models both source and user heterogeneity. Finally, we demonstrate the practical versatility of our framework by applying it to two downstream tasks.

The contributions of this paper are twofold. First, we propose a simple yet effective architecture that jointly leverages current and historical multi-channel wearable signals under missing and non-uniform feature patterns, and we systematically validate its design on two datasets. Second, we construct and will publicly release \textsc{ParroTao}, a large-scale, multi-sport, multi-device dataset that preserves device-specific feature sets and cross-user variation, offering a realistic testbed for modeling heterogeneity in real-world wearable data.

The rest of the paper is organized as follows. Section~\ref{sec:formulation} formulates the problem. Section~\ref{sec:method} elaborates on our method. Section~\ref{sec:experiments} presents our dataset, validates our approach, and provides two application examples. 
Finally, Section~\ref{sec:conclusion} provides conclusion. Source code can be found in supplementary materials.

\section{Problem Formulation} \label{sec:formulation}

\begin{figure*}[!t]
    \centering
    \includegraphics[width=\textwidth]
    {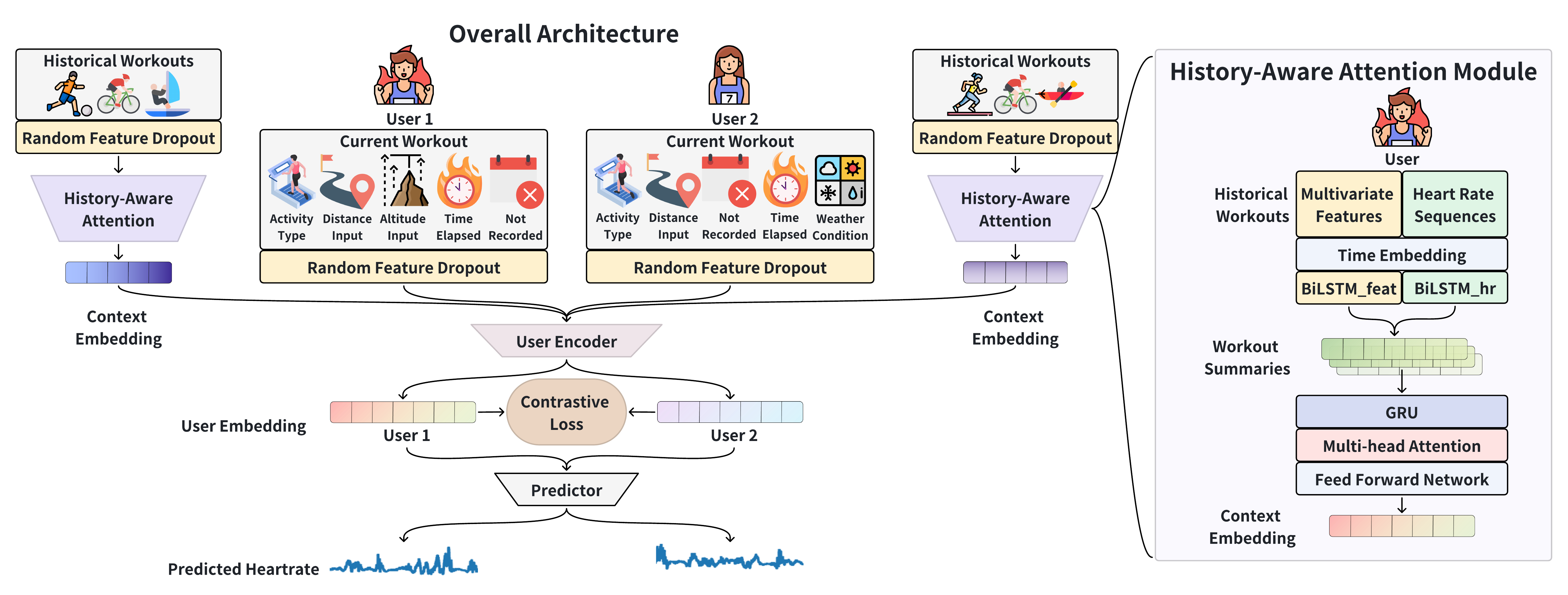}
    \caption{Architecture of the proposed framework. Random feature dropout acts on both historical and current inputs to alleviate source heterogeneity. A history-aware attention module compresses the historical record $\mathcal{H}_u$ into a context embedding $\mathbf{u}_u$, which is concatenated with the feature matrix of the current workout plan $\mathbf{X}^{(\mathrm{cur})}$ and fed to a user encoder. Finally, a joint objective combines mean-squared error with an InfoNCE contrastive loss that aligns semantically similar embeddings.}
    \label{fig:pipeline}
\end{figure*}

Our goal is to predict a user's heart rate sequence for a given workout session based on their historical data and the characteristics of the current session. A workout session refers to a continuous period of exercise, such as a 30-minute run or a one-hour cycling trip. This task mirrors a practical scenario where, for instance, a runner plans a route with a specific elevation profile and target pace and uses the model to forecast their physiological response, helping them select a suitable training plan.

We represent each workout session as a multivariate time series. For a user $u$, their history $\mathcal{H}_u$ consists of their past $N_u$ workout sessions:
\begin{equation}
    \label{eq:session}
    \mathcal{H}_u = \{ (\mathbf{X}^{(i)}, \mathbf{y}^{(i)}, \mathbf{a}^{(i)}, \Delta\tau^{(i)}) \}_{i=1}^{N_u}.
\end{equation}
Here, for the $i$-th historical session:
    $\mathbf{X}^{(i)} = (\mathbf{x}_1^{(i)}, \dots, \mathbf{x}_{T_i}^{(i)})$ is the sequence of multivariate features read from one sensor, where $T_i$ is the duration of the session. Each vector $\mathbf{x}_t^{(i)} \in \mathbb{R}^D$ contains sensor data at time $t$, such as pace, elevation, and cadence. We consider a global set of $D$ possible feature dimensions across all devices.
    $\mathbf{y}^{(i)} = (y_1^{(i)}, \dots, y_{T_i}^{(i)})$ is the corresponding sequence of observed heart rate values.
    $\mathbf{a}^{(i)} \in \mathbb{R}^m$ is a vector of static attributes for the session, such as activity type (e.g., running, cycling) and user ID.
    $\Delta\tau^{(i)}$ is the time interval between the end of session $i-1$ and the start of session $i$, capturing the temporal gap between workouts.
Due to the diversity of wearable devices, a session from a particular device may only provide a subset of the $D$ global features. For any feature dimension $d \in \{1, \dots, D\}$ that is unobserved in a session, its value is set to zero.

The prediction task is defined as follows: given a user's history $\mathcal{H}_u$ and the features of current planned training workout, $\mathbf{X}^{(\mathrm{cur})} \in \mathbb{R}^{D \times T}$, we aim to predict the heart rate sequence $\hat{\mathbf{y}}^{(\mathrm{cur})} = (\hat{y}_1, \dots, \hat{y}_T)$. This is achieved by learning a model $f_\theta$ with parameters $\theta$:
\begin{equation*}
    \hat{\mathbf{y}}^{(\mathrm{cur})} = f_\theta(\mathbf{X}^{(\mathrm{cur})}, \mathcal{H}_u).
\end{equation*}
The model parameters $\theta$ are optimized by minimizing a loss function over a training set of $N$ examples. Each sample consists of a target session and its corresponding history. The objective is to minimize the mean squared error (MSE) between the predicted and true heart rate sequences:
\begin{equation*}
    \mathcal{L}(\theta) = \frac{1}{N} \sum_{j=1}^{N} \frac{1}{T_j} \sum_{t=1}^{T_j} (\hat{y}_t^{(j)} - y_t^{(j)})^2.
\end{equation*}
This formulation explicitly addresses the challenge of data heterogeneity, distinguishing our approach from traditional ones that often hold the following simplified assumptions:

\textbf{(1) Source Heterogeneity:} 
    Traditional methods typically assume a fixed and complete feature space, meaning that every input $\mathbf{X}$ has the same set of features. Our formulation relaxes this by defining a global feature set of dimension $D$, where any given session's input $\mathbf{X}^{(i)}$ may contain only a sparse subset of these features. The model $f_\theta$ must therefore learn to be robust to varied feature availability across different devices and data sources.
    
\textbf{(2) User Heterogeneity:} Most existing methods implicitly assume that the data-generating process $P(\mathbf{y} | \mathbf{X}, u)$ remains stationary for each user, regardless of their training history. In contrast, our model conditions the predictive distribution on the user's historical data, and current activity type, i.e., $P(\mathbf{y}^{(\mathrm{cur})} | \mathbf{X}^{(\mathrm{cur})}, \mathcal{H}_u)$. This allows our model to capture user-specific and context-dependent variations in the relationship between sensor inputs and heart rate. As a result, the model learns a user-aware and context-sensitive mapping.


\section{Method}\label{sec:method}

To jointly address both \emph{source} heterogeneity and \emph{user} heterogeneity, we design a framework that learns a \emph{unified representation space} for heterogeneous heart rate time series. As illustrated in Fig.~\ref{fig:pipeline}, our framework adopts a multi-stage architecture that explicitly incorporates solutions for data heterogeneity into the representation learning process, resulting in robust user embeddings for precise heart rate prediction.

The process begins by applying a \emph{random-feature dropout} module to both the historical and current workout plan across different devices to ensure device-agnostic learning. Subsequently, the model operates in two main encoding stages. First, a \emph{history-aware attention module} processes the user's workout history $\mathcal{H}_u$ to generate a \textbf{context embedding}, $\mathbf{u}_u$, which encapsulates long-term physiological traits. Second, this context embedding is concatenated with the features of the current workout plan, $\mathbf{X}^{(\mathrm{cur})}$, and fed into a \emph{user encoder} to produce a final \textbf{user embedding}, $\mathbf{z}_u$. This final embedding, which integrates historical context with current session dynamics, is the cornerstone of our approach. It is regularized using a \emph{contrastive loss} to enforce physiological consistency and passed to a predictor to forecast the heart rate sequence.

\subsection{Random Feature Dropout}\label{subsec:dropout}
To mitigate source heterogeneity and force the model to learn from diverse feature subsets, we design a random feature dropout module that operates on the feature dimension. This is applied during training to all input features, i.e., both the current session's $\mathbf{X}^{(\mathrm{cur})}$ and all historical $\mathbf{X}^{(i)}$.

For any given feature matrix $\mathbf{X}^{(j)} \in \mathbb{R}^{D \times T_j}$ (where $j$ can be ``cur" or a historical index $i$), we generate a sample-specific binary mask $\mathbf{m}^{(j)} \in \{0, 1\}^D$. This mask determines which feature channels are kept ($m_{d}^{(j)}=1$) or dropped ($m_{d}^{(j)}=0$) for the entire time sequence $T_j$.
The probability $p$ of dropping a feature follows a curriculum strategy~\cite{bengio2009curriculum}. It gradually increases from a minimum value $p_{\min}$ to a maximum $p_{\max}$ over the first $E$ epochs, governed by the equation:
\[
p(e) = p_{\min} + (p_{\max} - p_{\min}) \cdot \min\left(\frac{e}{E}, 1.0\right),
\]
where $e$ is the current epoch number. This strategy allows the model to first learn from richer data before adapting to more challenging, sparse inputs.

To ensure model stability, we enforce two constraints. First, a predefined set of essential feature can be designated as ``main features", which are always protected from being dropped. During implementation, speed and altitude are designated as the main features, given their availability across all sensors. Second, it is guaranteed that at least $K$ features are retained for every sample. This dropout strategy ensures that the model does not become reliant on any specific sensor's feature set, thereby enhancing its robustness and generalizability.

\subsection{History-Aware Attention Module for Context Embedding}\label{subsec:history_encoder}

Fig.~\ref{fig:pipeline} depicts the proposed History-Aware Attention Module. The user's workout history, defined in Eq.~\ref{eq:session} , provides essential information about their baseline fitness and long-term trends. We distill this history into a fixed-size context embedding $\mathbf{u}_u$. The dynamic components ($\mathbf{X}^{(i)}$, $\mathbf{y}^{(i)}$, $\Delta\tau^{(i)}$) are processed by a hierarchical encoder, while the static attributes $\mathbf{a}^{(i)}$ (e.g., user ID, activity type) are utilized later for contrastive learning (see Section~\ref{subsec:contrastive}).

\textbf{1) Intra-Workout Encoding.} We first model the temporal dynamics within each historical workout. The time interval $\Delta\tau^{(i)}$ is mapped to a learned time embedding $\mathbf{e}_{\tau}^{(i)} \in \mathbb{R}^{D_t}$. This embedding is shared by all time steps and concatenated with both the feature and heart rate sequences. Two separate Bi-directional LSTMs (BiLSTMs)~\cite{hochreiter1997long} then encode these augmented sequences:
$$
\mathbf{w}_{\text{feat}}^{(i)} = \text{BiLSTM}_{\text{feat}}\!\left(\left[\mathbf{X}^{(i)}; \mathbf{e}_{\tau}^{(i)}\mathbf{1}^{\top}\right]\right), 
$$
$$
\mathbf{w}_{\text{hr}}^{(i)} = \text{BiLSTM}_{\text{hr}}\!\left(\left[\mathbf{y}^{(i)}; \mathbf{e}_{\tau}^{(i)}\mathbf{1}^{\top}\right]\right).
$$
The final hidden states of the BiLSTMs are concatenated to form a summary vector for each workout: $\mathbf{w}^{(i)} = [\mathbf{w}_{\text{feat}}^{(i)}; \mathbf{w}_{\text{hr}}^{(i)}] \in \mathbb{R}^{2H}$.

\textbf{2) Inter-Workout Modeling.} To capture the user's physiological evolution, the sequence of workout summaries $\{\mathbf{w}^{(i)}\}_{i=1}^{N_u}$ is processed by a Gated Recurrent Unit (GRU)~\cite{cho2014learning}:
\[
\{\mathbf{c}^{(i)}\}_{i=1}^{N_u} = \text{GRU}\left(\{\mathbf{w}^{(i)}\}_{i=1}^{N_u}\right).
\]
This produces a sequence of context vectors $\mathbf{c}^{(i)} \in \mathbb{R}^{H}$, each conditioned on all preceding workouts.

\textbf{3) Attention and Fusion.} Finally, a multi-head attention mechanism identifies the most relevant historical sessions. The attention output $\mathbf{a}$ is computed using the context vector of the most recent historical workout $\mathbf{c}^{(N_u)}$ as the query, and the full sequence $\{\mathbf{c}^{(i)}\}_{i=1}^{N_u}$ as keys and values, and the final \textbf{context embedding} $\mathbf{u}_u \in \mathbb{R}^{H_c}$ is produced by fusing the most recent context vector with the attention output via a feed-forward network (FFN):
$$
\mathbf{a} = \text{Attention}\left(\text{query}=\mathbf{c}^{(N_u)}, \text{key/value}=\{\mathbf{c}^{(i)}\}_{i=1}^{N_u}\right),
$$

$$
\mathbf{u}_u = \text{FFN}\left(\left[\mathbf{c}^{(N_u)}; \mathbf{a}\right]\right).
$$

\subsection{Contrastive Representation Learning}\label{subsec:contrastive}
While $\mathbf{u}_u$ captures historical patterns, the final representation must also incorporate the current workout plan. We broadcast $\mathbf{u}_u$ along the time dimension of the current workout plan's feature matrix, $\mathbf{X}^{(\mathrm{cur})} \in \mathbb{R}^{D \times T}$, and concatenate them to form an enriched input $[\mathbf{X}^{(\mathrm{cur})}; \mathbf{u}_u\mathbf{1}^{\top}] \in \mathbb{R}^{(D+H_c) \times T}$.
This combined representation is fed into a \textbf{user encoder}. The output of this encoder is taken as the definitive \textbf{user embedding}, $\mathbf{z}_u \in \mathbb{R}^{H_{z}}$. This embedding is a holistic representation, conditioned on both long-term history and immediate task characteristics.

To ensure the user embeddings $\mathbf{z}_u$ are physiologically meaningful, we regularize the embedding space using an InfoNCE contrastive loss~\cite{oord2018representation}. For a mini-batch of $B$ samples, we form pairs of embeddings and their group labels $\{(\mathbf{z}_b, g_b)\}_{b=1}^{B}$. The group label $g_b$ is derived from the static attributes $\mathbf{a}^{(\text{cur})}_b$ of the corresponding sample (e.g., user ID or activity type). We define $\mathcal{P}$ as the set of all positive pairs of indices. The loss is:
$$
\mathcal{L}_{\text{CL}}
\;=\;
-\frac{1}{|\mathcal{P}|}\sum_{(b,c)\in\mathcal{P}}
\log\frac{\exp (\mathbf{z}_b^{\top}\mathbf{z}_c/\tau)}
{\sum_{k=1, k\neq b}^{B}\exp (\mathbf{z}_b^{\top}\mathbf{z}_k/\tau)},
$$
$$
\mathcal{P}=\{(b,c)\mid g_b=g_c,\;b\neq c\}.
$$
This loss, applied after $\ell_2$ normalization with temperature $\tau$, pulls embeddings from the same group together, enforcing a consistent structure on the representation space.

\subsection{Training Objective}\label{subsec:loss}
The final user embedding $\mathbf{z}_u$ is passed to a feed-forward predictor network to generate the heart rate forecast, $\hat{\mathbf{y}}$. The model is trained end-to-end by minimizing a combined loss function over a mini-batch of $B$ samples:
\[
\mathcal{L}
\;=\;
\frac{1}{B}\sum_{b=1}^{B}\underbrace{\left(\frac{1}{T_b}\sum_{t=1}^{T_b}(y_{t}^{(\mathrm{b})}-\hat{y}_{t}^{(\mathrm{b})})^2\right)}_{\mathcal{L}_{\text{MSE}}^{(b)}}
\;+\;
\lambda\,\mathcal{L}_{\text{CL}},
\]
where $\mathcal{L}_{\text{MSE}}^{(b)}$ is the mean squared error for the $b$-th sample of duration $T_b$, and $\mathcal{L}_{\text{CL}}$ is the contrastive loss. The hyperparameter $\lambda$ balances point-wise accuracy against representation coherence.

\section{Experiments} \label{sec:experiments}

To comprehensively evaluate our proposed method, we design four corresponding groups of experiments: \textbf{(Group 1)} We evaluate the overall and per-sport predictive performance of our model against 8 baselines on the FitRec and \textsc{ParroTao} datasets. \textbf{(Group 2)} We conduct comprehensive ablation studies on both datasets to analyze the individual impact of our proposed modules. \textbf{(Group 3)} We qualitatively and quantitatively assess the discriminative power of the learned user embeddings. \textbf{(Group 4)} We demonstrate the model's utility in route recommendation and heart rate imputation.

\subsection{Dataset Description}\label{subsec:data}

To comprehensively evaluate our model, we utilize two distinct datasets. We first use FitRec, a widely used open-source benchmark, to assess performance under large-scale and relatively uniform conditions. To further examine our model’s robustness in the presence of real-world heterogeneity, we introduce \textsc{ParroTao}—a device-diverse dataset collected from multiple wearable brands.

\textbf{FitRec.}
The FitRec corpus, released by~\cite{ni2019modeling}, contains $167{,}373$ workout sessions from $956$ users across $40$ sport categories, scraped from the public platform \emph{Endomondo}. Each session is a multivariate time series recording core metrics alongside contextual metadata. As raw speed values can be unreliable, we follow the original study and use a \emph{derived speed}, computed from distance and time intervals. For modeling, each session is partitioned into non-overlapping windows of $450$ time steps. This results in an average of over $175$ windows per user, spanning approximately two years.

\textbf{ParroTao.}
To address the absence of publicly available, device-diverse corpora, we developed \textsc{ParroTao}, a comprehensive dataset containing 42,576 workout sessions from 113 recreational athletes. The data were collected using three leading wearable brands (\emph{Coros}, \emph{Garmin}, and \emph{Huawei}) over a three-year period from 2022 to 2025. The demographic and fitness characteristics of the \textsc{ParroTao} cohort are detailed in Table~\ref{tab:demographics}. \textsc{ParroTao} is intentionally heterogeneous, with partially overlapping sensor suites and different data units across vendors, reflecting real-world conditions.

Table~\ref{tab:variables_combined_small} provides a comprehensive, side-by-side comparison of the variables available in both datasets. This setup allows us to test our method's robustness against both relatively uniform data (FitRec) and highly heterogeneous data (\textsc{ParroTao}). More details about the datasets can be found in 
Supplementary Table I.

\begin{table}[!t]
    \centering
    \caption{Demographic and Fitness Characteristics of the \textsc{ParroTao} Dataset.}
    \label{tab:demographics}
    \begin{tabular}{lrr}
        \toprule
        \textbf{Characteristic} & \textbf{Count (Athletes)} & \textbf{Percentage} \\
        \midrule
        \multicolumn{3}{l}{\textit{Age Distribution}} \\
        \quad 18-20 years & 30 & 26.5\% \\
        \quad 21-23 years & 37 & 32.7\% \\
        \quad 24-26 years & 26 & 23.0\% \\
        \quad 27-30 years & 20 & 17.7\% \\
        \midrule
        \multicolumn{3}{l}{\textit{Gender Distribution}} \\
        \quad Male   & 79 & 69.9\% \\
        \quad Female & 34 & 30.1\% \\
        \midrule
        \multicolumn{3}{l}{\textit{Fitness Level Distribution}} \\
        \quad Amateur           & 52 & 46.0\% \\
        \quad Semi-Professional & 39 & 34.5\% \\
        \quad Professional      & 22 & 19.5\% \\
        \midrule
        \textbf{Total Athletes} & \textbf{113} & \textbf{100.0\%} \\
        \bottomrule
    \end{tabular}
\end{table}

\begin{table}
\centering
{
\setlength{\tabcolsep}{1pt}

\caption{Counts of recorded feature categories in the FitRec and \textsc{ParroTao} datasets.}
\label{tab:variables_combined_small}
\small

\begin{tabular}{lcccc}
\toprule
\textbf{Channel} & \textbf{FitRec} & \multicolumn{3}{c}{\textbf{ParroTao}} \\
\cmidrule(lr){3-5}
&  & \textbf{Coros} & \textbf{Garmin} & \textbf{Huawei} \\
\midrule
Core Spatiotemporal       & 3 & 5 & 5 & 3 \\
Cardiorespiratory & 1 & 3 & 6 & 4 \\
Running dynamics & 0 & 2 & 7 & 3 \\
Sport-specific extras     & 0 & 1 & 2 & 1 \\
Contextual Metadata       & 3 & 5 & 5 & 5 \\
\bottomrule
\end{tabular}
}
\end{table}

\subsection{Baselines and Implementations} \label{subsec:baseline}
\textbf{Baselines.} We compare our method with representative baselines from two broad families: physiology-grounded models and data-driven models. In addition, we include a simple User mean baseline to quantify the benefit of personalization. We followed the original implementation of the baseline for experiment.

\begin{itemize}
    \item \textbf{User mean.} For each user $u$, we predict the target variable at every time step by the mean value of that variable computed over all training sequences of $u$.
    \item \textbf{Multilayer Perceptron (MLP)~\cite{popescu2009multilayer}.} A feed-forward network with two hidden layers. At each time step, it takes the concatenated input features (e.g., kinematic signals, user embeddings, and contextual variables) and outputs the corresponding prediction.
    \item \textbf{Smartphone VO\(_2\)-Driven model~\cite{sumida2013estimating}.} A physiology-grounded baseline that computes an oxygen-demand proxy from resting metabolism, speed, and grade, and personalizes it using a user-specific scaling factor and resting heart rate normalization. The inputs are then mapped to heart rate using a feed-forward predictor.
    \item \textbf{Hybrid ODE–neural model~\cite{modeling2023personalized}.} A dynamical baseline in which heart rate evolution is described by an ordinary differential equation (ODE) system. The ODE parameters are conditioned on a learned user embedding obtained from the user's historical workouts. The personalized ODE is solved to predict the heart rate trajectory.
    \item \textbf{FitRec~\cite{ni2019modeling}.} A sequential model that uses user embeddings and recurrent encoders to jointly model user state and session dynamics. We adapt it to map input sequences and user IDs to per-timestep heart rate predictions.
    \item \textbf{STM-BiLSTM-Att~\cite{lin2023new}.} An attention-augmented recurrent architecture. It utilizes an LSTM layer for long-range temporal feature extraction, a BiLSTM layer to capture both forward and backward dependencies, and a final attention mechanism to re-weight the BiLSTM hidden states before making a prediction. 
    \item \textbf{Temporal Convolutional Network (TCN)~\cite{lea2017temporal}.} A Temporal Convolutional Network that applies dilated 1D convolutions over the sequence of concatenated features. Its residual blocks are designed to capture temporal patterns across a large receptive field to make per-timestep predictions.
    \item \textbf{Transformer~\cite{vaswani2017attention}.} A multi-layer Transformer encoder that processes the full input sequence. It uses self-attention mechanisms to model complex, long-range dependencies between all time steps, followed by a position-wise head to generate the output sequence.
\end{itemize}

\textbf{Architecture.} Our model consists of a backbone encoder for the current workout and a history encoder for user context. The current-session backbone is a two-layer LSTM with 128-dimensional hidden states, followed by dropout ($p{=}0.2$) and a GELU activation. To incorporate historical information, we employ a history-aware attention module over the $10$ most recent workouts of each user, comprising (i) time encoding, (ii) a 64-dimensional BiLSTM, (iii) a 128-dimensional cross-session GRU, and (iv) 4-head multi-head attention (hidden dimension 128). Categorical attributes (user ID, sport category, and gender) are represented as learned embeddings. The backbone input is the concatenation of per-timestep workout features, the history-encoder output, and the relevant context embeddings.

\textbf{Training Protocol.} We split the dataset into 80\%/10\%/10\% for training/validation/testing at the participant level: all sessions from a user are assigned to exactly one split to prevent identity leakage. Because users contribute varying numbers of sessions, we construct splits such that the total number of workouts is approximately in an 8:1:1 ratio across train/validation/test. We train using RMSProp (batch size 64, learning rate $1\times10^{-2}$), apply gradient clipping with threshold 2.0, and use early stopping with a patience of 10 epochs, selecting the checkpoint with the best validation performance. We include a contrastive objective with weight 0.1 and cosine-similarity temperature 0.1. 

\textbf{Evaluation.} We report Mean Squared Error (MSE) and Mean Absolute Error (MAE) on the held-out test set. To quantify uncertainty, we perform 200 bootstrap iterations, each sampling 80\% of the test set with replacement, and report the mean and standard deviation of each metric across iterations. Statistical significance is assessed using one-sided Wilcoxon signed-rank tests for paired comparisons and the Friedman test for multi-method comparisons.

\begin{table*}[!t]
\centering
{
\small
\caption{Overall performance comparison on the FitRec and \textsc{ParroTao} datasets. We report the Mean ± Std for MSE and MAE estimated over 200 bootstraps. The best results are highlighted in \textbf{bold}. Lower values indicate better performance.}
\label{tab:overall_performance}
\begin{tabular}{lcccc}
\toprule
\multicolumn{1}{c}{\multirow{2}{*}{\textbf{Model}}} & \multicolumn{2}{c}{\textbf{FitRec}} & \multicolumn{2}{c}{\textbf{ParroTao}} \\
\cmidrule(lr){2-3} \cmidrule(lr){4-5}
\multicolumn{1}{c}{} & \textbf{MSE} $\downarrow$ & \textbf{MAE} $\downarrow$ & \textbf{MSE} $\downarrow$ & \textbf{MAE} $\downarrow$ \\
\midrule
User Mean & 570.94 $\pm$ 10.42 & 18.06 $\pm$ 0.10 & 629.89 $\pm$ 17.26 & 19.82 $\pm$ 0.27 \\
\midrule
Smartphone VO$_2$-Driven & 355.62 $\pm$ 6.21 & 14.03 $\pm$ 0.07 & 311.53 $\pm$ 8.11 & 13.08 $\pm$ 0.16 \\
Hybrid ODE–Neural & 346.31 $\pm$ 5.82 & 13.81 $\pm$ 0.07 & 387.24 $\pm$ 9.18 & 15.27 $\pm$ 0.18 \\
\midrule
MLP & 370.26 $\pm$ 6.00 & 14.35 $\pm$ 0.07 & 284.62 $\pm$ 7.34 & 12.36 $\pm$ 0.16 \\
STM-BiLSTM-Att & 278.20 $\pm$ 3.06 & 12.50 $\pm$ 0.06 & 174.92 $\pm$ 6.07 & 9.70 $\pm$ 0.15 \\
FitRec & 248.02 $\pm$ 3.53 & 11.69 $\pm$ 0.06 & 139.53 $\pm$ 4.86 & 8.24 $\pm$ 0.11 \\
TCN & 254.93 $\pm$ 3.37 & 11.91 $\pm$ 0.06 & 160.36 $\pm$ 5.31 & 8.61 $\pm$ 0.13 \\
Transformer & 253.70 $\pm$ 3.16 & 11.79 $\pm$ 0.05 & 148.37 $\pm$ 5.36 & 8.24 $\pm$ 0.13 \\
\midrule
\textbf{Ours} & \textbf{204.63 $\pm$ 3.41} & \textbf{10.28 $\pm$ 0.05} & \textbf{125.04 $\pm$ 4.57} & \textbf{7.53 $\pm$ 0.11} \\
\bottomrule
\end{tabular}
}
\end{table*}

\begin{table}[!t]
\centering
\caption{Statistical analysis of our model against baselines using a one-sided Wilcoxon signed-rank test. We report the relative improvement (Imp., \%) and Cohen's $d$ effect size. For all comparisons, the FDR-corrected p-value ($p_{adj}$) is less than 0.001, indicating statistically significant improvements.}
\label{tab:statistical_analysis}

\setlength{\tabcolsep}{4.2pt}
\renewcommand{\arraystretch}{1.05}
\small

\resizebox{\columnwidth}{!}{%
\begin{tabular}{lcc}
\toprule
\textbf{Baseline Compared} & \textbf{MSE (Imp.\% $\uparrow$ / $d$)} & \textbf{MAE (Imp.\% $\uparrow$ / $d$)} \\
\midrule
\multicolumn{3}{c}{\textbf{FitRec Dataset}} \\
\midrule
User Mean                & 64.16 / -35.06 & 43.05 / -81.26 \\
MLP                      & 44.73 / -28.11 & 28.35 / -65.01 \\
Smartphone VO$_2$-Driven & 42.46 / -24.81 & 26.70 / -60.82 \\
Hybrid ODE--Neural       & 40.91 / -24.84 & 25.50 / -56.15 \\
STM-BiLSTM-Att           & 26.44 / -28.87 & 17.70 / -48.57 \\
FitRec                   & 17.49 / -12.53 & 12.00 / -32.18 \\
TCN                      & 19.73 / -14.38 & 13.66 / -35.31 \\
Transformer              & 19.34 / -15.72 & 12.79 / -36.00 \\
\midrule
\multicolumn{3}{c}{\textbf{ParroTao Dataset}} \\
\midrule
User Mean                & 80.15 / -30.62 & 62.04 / -48.38 \\
MLP                      & 56.07 / -25.28 & 39.09 / -35.94 \\
Smartphone VO$_2$-Driven & 59.86 / -26.75 & 42.49 / -40.09 \\
Hybrid ODE--Neural       & 67.71 / -30.12 & 50.73 / -43.30 \\
STM-BiLSTM-Att           & 28.52 / -10.48 & 22.43 / -17.98 \\
FitRec                   & 10.38 / -5.21  & 8.65 / -11.05 \\
TCN                      & 22.03 / -10.73 & 12.64 / -13.79 \\
Transformer              & 15.73 / -5.81  & 8.70 / -7.99 \\
\bottomrule
\end{tabular}%
}
\end{table}

\subsection{Group 1: Predictive Performance} \label{subsec:g1}
\textbf{Overall Performance.} Table~\ref{tab:overall_performance} compares our model with the baselines on FitRec and \textsc{ParroTao}. Our approach achieves the lowest MSE and MAE on both datasets, reducing MSE by 17.5\% on FitRec and 10.4\% on \textsc{ParroTao} relative to the strongest competitors. We use the one-sided Wilcoxon signed-rank test to compare our model with each baseline, applying the Benjamini–Hochberg FDR correction ($q=0.05$) for multiple comparisons. For each baseline, Wilcoxon signed rank tests consistently yield $p < 0.001$ with absolute Cohen's $d > 6$, demonstrating that our improvements are statistically significant and substantial. Detailed statistics are reported in Table~\ref{tab:statistical_analysis}.

\textbf{Per-Sport Analysis.} To provide a granular assessment of our model's capabilities, we evaluated its performance on individual sport types. Our model consistently achieves top-1 performance across individual sports on both FitRec and \textsc{ParroTao} datasets, obtaining 13 first-place rankings for MSE on FitRec and 6 on \textsc{ParroTao} as shown in Table~\ref{tab:number-1-rank}. The model achieves the lowest MSE and MAE for prevalent activities such as running and cycling. It also demonstrates remarkable robustness in data-scarce scenarios (e.g. lowest MSE of 16.28 for Elliptical training despite only 153 samples). Detailed per-sport results can be found in 
Supplementary Tabels II and III. 

We use the Friedman test to compare model ranks across sports, followed by pairwise Wilcoxon tests with FDR correction. We also report Cohen's d to quantify effect sizes. As shown in Tables~\ref{tab:stat_ranks}and~\ref{tab:stat_wilcoxon}, statistical validation confirms our model significantly outperforms all baselines across fine-grained sport categories on both datasets. 

\begingroup
\setlength{\tabcolsep}{5pt}
\renewcommand{\arraystretch}{1.05}

\begin{table}[!t]
\centering
\caption{Number of first-place rankings in MSE and MAE across models on the FitRec and \textsc{ParroTao} datasets. Values are reported as (MSE, MAE).}
\label{tab:number-1-rank}
\small

\begin{tabular}{lcc}
\toprule
\textbf{Model} & \textbf{FitRec (MSE, MAE)} & \textbf{\textsc{ParroTao} (MSE, MAE)} \\
\midrule
Mean    & (2, 2)   & (0, 0) \\
Smart.  & (0, 1)   & (1, 1) \\
Hybrid. & (0, 0)   & (0, 0) \\
MLP     & (1, 1)   & (1, 1) \\
STM.    & (1, 1)   & (0, 0) \\
TCN     & (2, 1)   & (0, 0) \\
FitRec  & (3, 3)   & (2, 1) \\
Trans.  & (3, 4)   & (5, 5) \\
Ours    & (13, 12) & (4, 5) \\
\bottomrule
\end{tabular}
\end{table}

\endgroup

\begin{table}[tbp]
\centering
{
\caption{Friedman test average ranks for each model on both datasets (lower is better). The p-values confirm a significant difference among models.}
\label{tab:stat_ranks}
\setlength{\tabcolsep}{1pt} 
\begin{tabular}{lcccc}
\toprule
\multirow{2}{*}{\textbf{Model}} & \multicolumn{2}{c}{\textbf{FitRec}} & \multicolumn{2}{c}{\textbf{ParroTao}} \\
\cmidrule(lr){2-3} \cmidrule(lr){4-5}
& \textbf{MSE Rank} & \textbf{MAE Rank} & \textbf{MSE Rank} & \textbf{MAE Rank} \\
\midrule
User Mean & 7.28 & 7.36 & 8.46 & 8.31 \\
Smartphone VO2-Driven & 5.76 & 5.76 & 6.00 & 5.92 \\
Hybrid ODE–Neural & 6.08 & 6.16 & 7.31 & 7.46 \\
MLP & 6.00 & 6.16 & 5.23 & 5.00 \\
STM-BiLSTM-Att & 4.96 & 5.04 & 4.00 & 4.23 \\
TCN & 4.12 & 4.00 & 3.54 & 3.46 \\
FitRec & 3.68 & 3.56 & 3.92 & 4.00 \\
Transformer & 4.32 & 4.20 & 3.38 & 3.38 \\
\textbf{Ours} & \textbf{2.80} & \textbf{2.76} & \textbf{3.15} & \textbf{3.23} \\
\midrule
\textbf{Friedman p-value} & 1.32e-08 & 3.55e-09 & 6.87e-08 & 1.10e-07 \\
\bottomrule
\end{tabular}
}
\end{table} 

\begin{table}[!t]
\centering
\caption{Pairwise Wilcoxon signed-rank test results of our model against baselines. We report the win-draw-loss (w-d-l) statistics.}
\label{tab:stat_wilcoxon}

\setlength{\tabcolsep}{4.2pt} 
\renewcommand{\arraystretch}{1.05} 
\small

\resizebox{\columnwidth}{!}{%
\begin{tabular}{llcc}
\toprule
\textbf{Dataset} & \textbf{Comparison (Ours vs.)} & \textbf{MSE (w-d-l)} & \textbf{MAE (w-d-l)} \\
\midrule
\multirow{8}{*}{FitRec}
& FitRec                & 17-0-8 & 17-0-8 \\
& User Mean             & 21-0-4 & 22-0-3 \\
& MLP                   & 20-0-5 & 21-0-4 \\
& TCN                   & 18-0-7 & 19-0-6 \\
& Transformer           & 19-0-6 & 17-0-8 \\
& STM-BiLSTM-Att        & 19-0-6 & 20-0-5 \\
& Hybrid ODE--Neural    & 21-0-4 & 21-0-4 \\
& Smartphone VO2-Driven & 20-0-5 & 19-0-6 \\
\midrule
\multirow{8}{*}{\textsc{ParroTao}}
& FitRec                & 9-0-4  & 9-0-4 \\
& Mean                  & 12-0-1 & 12-0-1 \\
& MLP                   & 9-0-4  & 9-0-4 \\
& TCN                   & 9-0-4  & 7-0-6 \\
& Transformer           & 7-0-6  & 7-0-6 \\
& STM-BiLSTM-Att        & 8-0-5  & 9-0-4 \\
& Hybrid ODE--Neural    & 12-0-1 & 12-0-1 \\
& Smartphone VO2-Driven & 10-0-3 & 10-0-3 \\
\bottomrule
\end{tabular}%
}
\end{table}

\subsection{Group 2: Ablation Studies} \label{subsec:g2}
\subsubsection{Experiment Setup}

To systematically validate the effectiveness of our proposed approach, we conducted comprehensive ablation studies at three aspects: module-level ablation, component analysis, and feature-level ablation.

\paragraph{Module-level ablation}
We evaluated the contribution of each major component by systematically removing key modules from the full model. Specifically, we examined three variants: (i) a model without random feature dropout (w/o Dropout), (ii) a model without the history-aware attention module (w/o HAT), and (iii) a model without contrastive loss (w/o Contra.).

\paragraph{Component analysis}
To further understand the design choices within critical modules, we conducted ablation experiments on three setups. 

\textbf{Structure of History-Aware Attention:} We investigated alternative architectures for the history-aware attention module by replacing either the GRU or BiLSTM component with other sequential encoders, including unidirectional LSTM and Transformer encoders. This allows us to determine whether performance gains stem from the specific encoder choice or from the overall module design. 

\textbf{Label Strategies for Contrastive Learning:} We analyzed the impact of different label types in contrastive learning by comparing three variants: using only user IDs as labels, using only activity types as labels, and using both jointly. This helps us understand how different supervision signals contribute to representation learning.

\textbf{Length of Input History:} We examined the effect of history length $K$ by evaluating multiple values $K \in \{1, 5, 10, 20, 30\}$ on the FitRec dataset. This analysis reveals the optimal temporal context window for capturing historical patterns.

\paragraph{Feature-level ablation}
We analyzed input feature importance at two levels. At the global level, we performed feature ablations on both FitRec and \textsc{ParroTao} datasets to quantify performance degradation when individual input channels are removed. At the granular level, we examined how feature importance varies across sports and users. This provides insights into which features are most critical for accurate prediction and how feature importance varies across different contexts.

\subsubsection{Results of Module-Level Ablation}

Table~\ref{tab:ablation_merged} presents the module-level ablation results on both FitRec and \textsc{ParroTao} datasets. The full model consistently achieves the lowest MSE and MAE across both datasets, demonstrating that each component contributes positively to overall performance. All corrected $p$-values from the one-sided Wilcoxon signed-rank tests are below $1.0 \times 10^{-5}$, confirming that each module makes a statistically significant contribution.

Removing contrastive learning produces the most substantial performance degradation, with MSE increasing by 20.3\% on FitRec and 35.9\% on \textsc{ParroTao}. This confirms the critical role of contrastive learning in producing discriminative representations that enable accurate prediction. Excluding the history-aware attention module also degrades performance considerably, particularly on the more challenging \textsc{ParroTao} dataset where MSE rises by 17.0\%. This validates the importance of modeling long-range temporal dependencies in workout sequences. Eliminating random feature dropout leads to smaller yet consistent increases in both MSE and MAE, highlighting its contribution as a regularizer that mitigates device-specific overfitting. Overall, these results underscore the synergy of our architectural design, where each module addresses a distinct challenge in cross-device workout performance prediction.

\begin{table}[!t]
\centering
\small
\setlength{\tabcolsep}{5pt}
\caption{Ablation results on FitRec and \textsc{ParroTao}. We report Mean $\pm$ Std for MSE and MAE estimated over 200 bootstraps. Cohen's $d$ (C's $d$) is computed by comparing the full model against each ablated variant on the same dataset and metric. ``w/o'' denotes ``without''. The best results within each dataset block are highlighted in \textbf{bold}. All corrected p-values from the one-sided Wilcoxon signed-rank tests satisfy $p_{\mathrm{adj}} < 1.0 \times 10^{-5}$.}
\label{tab:ablation_merged}
\begin{tabular}{lcc|cc}
\toprule
\textbf{Variant} & \textbf{MSE $\downarrow$} & \textbf{C's  $d$} & \textbf{MAE $\downarrow$} & \textbf{C's $d$} \\
\midrule
\multicolumn{5}{l}{\textbf{FitRec}} \\
\midrule
Ours (Full)     & \textbf{204.63 $\pm$ 3.41} & --     & \textbf{10.28 $\pm$ 0.05} & -- \\
w/o Dropout     & 213.40 $\pm$ 3.15          & -2.963 & 10.62 $\pm$ 0.05          & -9.956 \\
w/o HAT         & 209.52 $\pm$ 2.93          & -1.800 & 10.61 $\pm$ 0.05          & -9.941 \\
w/o Contra. & 246.08 $\pm$ 4.33          & -9.481 & 11.50 $\pm$ 0.06          & -26.001 \\
\midrule
\multicolumn{5}{l}{\textbf{\textsc{ParroTao}}} \\
\midrule
Ours (Full)     & \textbf{125.04 $\pm$ 4.57} & --     & \textbf{7.53 $\pm$ 0.11}  & -- \\
w/o Dropout     & 127.42 $\pm$ 5.04          & -0.332 & 7.75 $\pm$ 0.11           & -1.318 \\
w/o HAT         & 144.34 $\pm$ 5.45          & -2.551 & 8.40 $\pm$ 0.13           & -4.896 \\
w/o Contra. & 142.48 $\pm$ 5.57          & -2.289 & 8.21 $\pm$ 0.13           & -3.829 \\
\bottomrule
\end{tabular}
\end{table}

\subsubsection{Results of Component Analysis}

\paragraph{Structure of History-Aware Attention}
We examined alternative sequential encoders within the history-aware attention module by replacing either the GRU or BiLSTM component with alternative encoders. Table~\ref{tab:encoder_ablation} shows that these substitutions lead to only small fluctuations in MSE and MAE across both datasets. The consistent performance across different encoder architectures demonstrates the robustness of our module design and indicates that the key contribution lies in the attention-based integration of historical information rather than in the specific implementation of temporal encoding.

\begin{table}[!t]
    \centering
    \caption{Encoder choice in the history-aware attention module.
    We report mean $\pm$ std of MSE and MAE on the FitRec and
    \textsc{ParroTao} datasets when replacing either the first (default: BiLSTM) or
    second (default: GRU) encoder slot with alternative architectures.}
    \label{tab:encoder_ablation}
    \begin{tabular}{llll}
        \toprule
        Dataset & Modification & MSE $\downarrow$ & MAE $\downarrow$ \\
        \midrule
        \textbf{FitRec} 
        & Original
        & \textbf{204.63} $\pm$ \textbf{3.41}  & 10.28 $\pm$ 0.05 \\
        & BiLSTM $\rightarrow$ GRU
        & 248.84 $\pm$ 3.16  & 11.78 $\pm$ 0.06 \\
        & BiLSTM $\rightarrow$ LSTM 
        & 206.20 $\pm$ 4.09 & \textbf{10.27} $\pm$ \textbf{0.06} \\
        & BiLSTM $\rightarrow$ Transformer 
        & 206.30 $\pm$ 3.48 & 10.30 $\pm$ 0.06 \\
        & GRU $\rightarrow$ BiLSTM 
        & 212.42 $\pm$ 3.49 & 10.56 $\pm$ 0.06 \\
        & GRU $\rightarrow$ LSTM 
        & 210.42 $\pm$ 2.88 & 10.63 $\pm$ 0.06 \\
        & GRU $\rightarrow$ Transformer 
        & 208.89 $\pm$ 3.50 & 10.34 $\pm$ 0.06 \\
        \midrule
        \textbf{ParroTao}
        & Original
        & 125.04 $\pm$ 4.57  & \textbf{7.53} $\pm$ \textbf{0.11} \\
        & BiLSTM $\rightarrow$ GRU 
        & \textbf{124.01} $\pm$ \textbf{4.49} & 7.63 $\pm$ 0.11 \\
        & BiLSTM $\rightarrow$ LSTM 
        & 128.53 $\pm$ 4.72 & 7.80 $\pm$ 0.12 \\
        & BiLSTM $\rightarrow$ Transformer 
        & 127.06 $\pm$ 4.84 & 7.77 $\pm$ 0.11 \\
        & GRU $\rightarrow$ BiLSTM 
        & 125.57 $\pm$ 4.79 & 7.65 $\pm$ 0.11 \\
        & GRU $\rightarrow$ LSTM 
        & 125.29 $\pm$ 4.86 & 7.54 $\pm$ 0.12 \\
        & GRU $\rightarrow$ Transformer 
        & 127.71 $\pm$ 4.80 & 7.79 $\pm$ 0.11 \\
        \bottomrule
    \end{tabular}
\end{table}

\paragraph{Label Strategies for Contrastive Learning}
Table~\ref{tab:contrastive_ablation} presents the results of three label strategies for contrastive learning. The joint-label setting consistently achieves the best performance on both FitRec and \textsc{ParroTao} datasets. User-ID supervision encourages the encoder to capture user-specific preferences and physiological patterns across sessions, while activity-type supervision promotes better separation between different sports. Combining both sources of supervision therefore leads to more accurate and robust downstream predictions.

\begin{table}[!t]

\centering
\caption{Ablation study of contrastive labels on FitRec and \textsc{ParroTao} datasets.}
\setlength{\tabcolsep}{2pt} 
\begin{tabular}{l l cc}
\toprule
Dataset & Contrastive labels & MSE (mean $\pm$ std) & MAE (mean $\pm$ std) \\
\midrule
\multirow{2}{*}{FitRec} 
& User ID only      & $211.16 \pm 3.44$ & $10.50 \pm 0.05$ \\
& Sport type only   & $227.79 \pm 3.26$ & $11.10 \pm 0.06$ \\
& User ID and Sport type   & $\textbf{204.63} \pm \textbf{3.41}$ & $\textbf{10.28} \pm \textbf{0.05}$ \\
\midrule
\multirow{2}{*}{ParroTao} 
& User ID only      & $128.62 \pm 4.72$ & $7.79 \pm 0.12$ \\
& Sport type only   & $136.79 \pm 4.98$ & $8.19 \pm 0.11$ \\
& User ID and Sport type   & $\textbf{125.04} \pm \textbf{4.57}$ & $\textbf{7.53} \pm \textbf{0.11}$ \\
\bottomrule
\end{tabular}
\label{tab:contrastive_ablation}
\end{table}

\paragraph{Length of Input History}
We evaluate history lengths $K \in \{1,5,10,20,30\}$ on the FitRec dataset and report overall MSE and MAE in Table~\ref{tab:fitrec_histlen_overall}. Moving from $K{=}1$ to $K{=}5$ yields almost no improvement in aggregate error. In contrast, increasing the history length to $K{=}10$ reduces MSE by approximately 3.3\% and MAE by 0.6\% compared to $K{=}5$. Using $K{=}20$ workouts achieves similar performance with slightly lower errors, while $K{=}30$ further improves MAE but begins to increase MSE again. Overall, the range $K \approx 10$--$20$ appears near-optimal at the dataset level, and $K{=}10$ offers a practical operating point that clearly outperforms short histories while avoiding the additional computational cost of longer windows.

To examine heterogeneity across sports and users, Table~\ref{tab:fitrec_histlen_best} reports how many sports and users achieve their lowest per-entity MSE at each history length. At the sport level, $8/25$ sports reach their best performance with a short history ($K{=}1$ or $5$), $3/25$ prefer $K{=}10$, and $14/25$ benefit from longer histories ($K{=}20$ or $30$). . t the user level, no single value of $K$ dominates: different users prefer short, intermediate, or long histories, and $K{=}10$ is competitive (it is the best setting for $22.0\%$ of users) without being extreme in either direction.

\begin{table}[!t]

  \centering
  \caption{Effect of history length $K$ (number of past workouts) on overall
  prediction error on the \textsc{FitRec} dataset. We report test MSE and MAE
  for each $K$. $\Delta$MSE and $\Delta$MAE denote the difference with respect
  to $K{=}10$ (negative values indicate better performance than $K{=}10$).}
  \label{tab:fitrec_histlen_overall}
  \setlength{\tabcolsep}{2pt} 
  \begin{tabular}{rcccc}
    \toprule
    $K$ & MSE & MAE & $\Delta$MSE vs.\ $K{=}10$ & $\Delta$MAE vs.\ $K{=}10$ \\
    \midrule
     1 & 206.47 & 10.52 &  +6.82 & +0.26 \\
     5 & 206.40 & 10.32 &  +6.75 & +0.06 \\
    10 & 199.65 & 10.26 &   0.00 &  0.00 \\
    20 & 199.50 & 10.24 &  -0.15 & -0.02 \\
    30 & 203.24 & 10.17 &  +3.59 & -0.09 \\
    \bottomrule
  \end{tabular}
\end{table}

\begin{table}[!t]

  \centering
  \caption{Number of sports and users for which a given history length $K$
  achieves the lowest per-entity MSE on \textsc{FitRec}. Percentages are
  relative to 25 sports and 91 users on the test set, respectively, and
  illustrate that no single history length is uniformly optimal across all
  entities.}
  \label{tab:fitrec_histlen_best}
  \begin{tabular}{rcccc}
    \toprule
    $K$ & \#sports best & \% sports & \#users best & \% users \\
    \midrule
     1  &  3 & 12.0\% &  7 &  7.69\% \\
     5  &  5 & 20.0\% & 23 & 25.3\% \\
    10  &  3 & 12.0\% & 20 & 22.0\% \\
    20  &  9 & 36.0\% & 17 & 18.7\% \\
    30  &  5 & 20.0\% & 24 & 26.4\% \\
    \bottomrule
  \end{tabular}
\end{table}

To understand how the model utilizes historical information, we further inspect the attention weights of the history-aware attention module when $K{=}10$. Table~\ref{tab:fitrec_attn_overall} shows that attention mass decays almost monotonically with workout age, indicating that the model emphasizes recent sessions rather than uniformly averaging over the entire history. Table~\ref{tab:fitrec_attn_sport_argmax} reveals sport-specific patterns: in 21 out of 25 sports (84\%), the maximum attention lies within the five most recent workouts, while only a minority peak at longer lags ($t_7$--$t_{10}$). This demonstrates that different sports exhibit distinct temporal dependence patterns, with most relying primarily on very recent sessions and a few placing more weight on longer-term history.

Taken together, these results suggest that: (i) using approximately ten past workouts is sufficient to capture most useful historical structure when averaged over the dataset, (ii) genuine variability exists across sports and users, with minorities favoring very short or very long histories, and (iii) $K{=}10$ strikes a reasonable balance between accuracy and computational cost while the attention mechanism adaptively emphasizes the most relevant historical information for each prediction.

\begin{table}[!t]

\centering
\caption{Average query attention weight over the ten most recent workouts on FitRec when the history length is $K{=}10$. ``Cum.'' reports the cumulative attention mass up to $t_k$.}
\label{tab:fitrec_attn_overall}

\scriptsize
\setlength{\tabcolsep}{2.5pt}

\begin{tabular}{lcccccccccc}
\toprule
 & $t_1$ & $t_2$ & $t_3$ & $t_4$ & $t_5$ & $t_6$ & $t_7$ & $t_8$ & $t_9$ & $t_{10}$ \\
\midrule
Mean weight & 0.207 & 0.135 & 0.115 & 0.097 & 0.085 & 0.075 & 0.073 & 0.072 & 0.071 & 0.070 \\
Cum.\ mass  & 0.207 & 0.342 & 0.456 & 0.553 & 0.638 & 0.713 & 0.786 & 0.858 & 0.929 & 1.000 \\
\bottomrule
\end{tabular}
\end{table}

\begin{table}[!t]

\centering
\caption{Location of the maximum query attention weight per sport on FitRec when using 10 past workouts ($K{=}10$). For each lag $t_k$, we report how many sports have their largest mean attention placed on $t_k$.}
\label{tab:fitrec_attn_sport_argmax}
\setlength{\tabcolsep}{2.8pt} 
\begin{tabular}{lcccccccccc}
\toprule
$\arg\max_k \text{ attention}(t_k)$ 
& $t_1$ & $t_2$ & $t_3$ & $t_4$ & $t_5$ & $t_6$ & $t_7$ & $t_8$ & $t_9$ & $t_{10}$ \\
\midrule
\# sports    & 13   & 1    & 3    & 2    & 2    & 0    & 1    & 1    & 0    & 2      \\
\% of sports & 52.0 & 4.0  & 12.0 & 8.0  & 8.0  & 0.0  & 4.0  & 4.0  & 0.0  & 8.0    \\
\bottomrule
\end{tabular}
\end{table}

\subsubsection{Results of Feature-Level Ablation}

\textbf{Evaluation on Feature Importance:}
Table~\ref{tab:fitrec_ablation_global} presents the global ablation results on FitRec, where the full model uses three sequential inputs: distance, altitude, and time elapsed. Removing distance leads to substantial performance degradation compared to the full model, while removing altitude has a smaller but still clearly detrimental effect. On the \textsc{ParroTao} dataset with 14 sequential input channels, Table~\ref{tab:parrotao_overall_ablation} shows that dropping most features increases both MSE and MAE. Power, cadence, and speed produce the largest performance degradation when ablated, indicating that they are globally the most influential features under our normalized representation.

\begin{table}[!t]

  \centering
  \caption{Global effect of dropping distance or altitude on FitRec.
  We report test MSE and MAE (mean $\pm$ standard deviation over users and sports)
  for the full model and its ablated variants.}
  \label{tab:fitrec_ablation_global}
  \begin{tabular}{llrrr}
    \toprule
    Model & Metric & Full model & w/o distance & w/o altitude \\
    \midrule
    Ours & MSE & \textbf{204.63} $\pm$ \textbf{3.41} & 522.11 $\pm$ 5.60 & 284.68 $\pm$ 3.96 \\
    Ours & MAE & \textbf{10.28} $\pm$ \textbf{0.05} & 17.67 $\pm$ 0.09 & 12.21 $\pm$ 0.07 \\
    \bottomrule
  \end{tabular}
\end{table}

\begin{table}[!t]

  \centering
  \caption{Overall effect of dropping each input feature on \textsc{ParroTao}. For each feature, we report the test error of the ablated model (``MSE$_\text{drop}$'' and ``MAE$_\text{drop}$'') and the change relative to the full model ($\Delta$MSE, $\Delta$MAE). Positive $\Delta$ indicates worse performance (higher error).}
  \label{tab:parrotao_overall_ablation}
  \begin{tabular}{lrrrr}
    \toprule
    Feature & MSE$_\text{drop}$ & $\Delta$MSE & MAE$_\text{drop}$ & $\Delta$MAE \\
    \midrule
    speed                & 164.13 &  39.09 &  8.76 &  1.24 \\
    cadence              & 172.30 &  47.25 &  9.09 &  1.56 \\
    power                & 312.26 & 187.22 & 13.89 &  6.36 \\
    stance time          & 127.20 &   2.16 &  7.70 &  0.17 \\
    temperature          & 136.30 &  11.26 &  7.72 &  0.20 \\
    enhanced altitude    & 130.60 &   5.56 &  7.68 &  0.16 \\
    position latitude    & 134.96 &   9.92 &  7.86 &  0.33 \\
    position longitude   & 128.25 &   3.21 &  7.63 &  0.10 \\
    step length          & 137.88 &  12.83 &  8.09 &  0.56 \\
    cycle length 16      & 124.70 &  -0.34 &  7.53 &  0.01 \\
    vertical oscillation & 125.67 &   0.62 &  7.58 &  0.05 \\
    vertical ratio       & 129.95 &   4.90 &  7.82 &  0.30 \\
    distance in meters   & 124.33 &  -0.72 &  7.52 & -0.01 \\
    elevation in meters  & 124.07 &  -0.97 &  7.51 & -0.01 \\
    \bottomrule
  \end{tabular}
\end{table}

\textbf{Heterogeneity across Sports and Users:}
Beyond these global averages, we also investigate how feature importance varies across sports and users. Feature importance exhibits substantial variability across different contexts. On FitRec, Table~\ref{tab:fitrec_ablation_hetero} shows that for 21 out of 25 sports (84.0\%), removing distance increases MSE more than removing altitude, while 3 out of 25 sports (12.0\%) are more sensitive to altitude. A similar pattern appears across users: for $72/91$ users ($79.1\%$), distance is more important, whereas for $16/91$ users ($18.6\%$) altitude is more critical. On \textsc{ParroTao}, Table~\ref{tab:parrotao_heterogeneity} reveals that power is the most critical feature for the majority of users, but cadence, speed, stance time, temperature, and distance each dominate for at least one user. These results demonstrate that the effective importance of each channel is strongly conditioned on sport type and user-specific characteristics.

\begin{table}[!t]

  \centering
  \caption{Heterogeneity of feature importance across sports and users on FitRec. ``Distance $>$ altitude'' counts sports/users for which dropping distance increases MSE more than dropping altitude, or where only the distance ablation increases MSE. Percentages are relative to 25 sports and 91 users on the test set.}
  \label{tab:fitrec_ablation_hetero}
  \begin{tabular}{lrr}
    \toprule
    Pattern & Sports & Users \\
    \midrule
    Distance $>$ altitude
      & $21$ ($84.0\%$) & $72$ ($79.1\%$) \\
    Altitude $>$ distance
      & $3$ ($12.0\%$)  & $16$ ($18.6\%$) \\
    Neither clearly important
      & $1$ ($4.0\%$)   & $3$ ($3.3\%$) \\
    \bottomrule
  \end{tabular}
\end{table}

\begin{table}[!t]

  \centering
  \caption{Heterogeneity of feature importance across sports and users on \textsc{ParroTao}.
  For each feature, we report the average change in MSE when the feature is
  dropped (Sport/User $\Delta$MSE) and the fraction of sports/users for which the
  ablation increases MSE ($\Delta$MSE$>0$). There are 13 sports and 19 users in the test set.}
  \label{tab:parrotao_heterogeneity}
  \scriptsize
\setlength{\tabcolsep}{6pt}
  \begin{tabular}{lrrrr}
    \toprule
  \multirow{2}{*}{Feature} & \multicolumn{2}{c}{Sport} & \multicolumn{2}{c}{User} \\
  \cmidrule(lr){2-3} \cmidrule(lr){4-5}
  & $\Delta$MSE & $\Delta$MSE$>0$ & $\Delta$MSE & $\Delta$MSE$>0$ \\
  \midrule
    speed                & 149.1 & 84.6\% &  47.8 & 94.7\% \\
    cadence              & 100.3 & 76.9\% &  88.1 &100.0\% \\
    power                & 166.4 & 84.6\% & 149.4 & 94.7\% \\
    stance time          &  -3.5 & 38.5\% &   7.4 & 31.6\% \\
    temperature          &   3.5 & 61.5\% &  13.3 & 57.9\% \\
    enhanced altitude    &   5.3 & 53.9\% &  13.8 & 73.7\% \\
    position latitude    &  31.6 & 61.5\% &   7.1 & 78.9\% \\
    position longitude   &   7.8 & 53.9\% &   1.1 & 57.9\% \\
    step length          &  13.7 & 69.2\% &  16.6 & 84.2\% \\
    cycle length 16      &  -0.8 & 38.5\% &   0.1 & 42.1\% \\
    vertical oscillation &   5.4 & 69.2\% &  -0.6 & 36.8\% \\
    vertical ratio       &  -2.3 & 38.5\% &   6.5 & 52.6\% \\
    distance in meters   &  -1.6 & 46.2\% &   8.8 & 52.6\% \\
    elevation in meters  &  -2.9 & 38.5\% &  -0.7 & 42.1\% \\
    \bottomrule
  \end{tabular}
\end{table}

\subsection{Group 3: Analysis of Learned Representations} \label{subsec:g3}

\begin{figure}[!t]
  \centering
  \begin{subfigure}[t]{\linewidth}
    \centering
    \includegraphics[width=\linewidth]{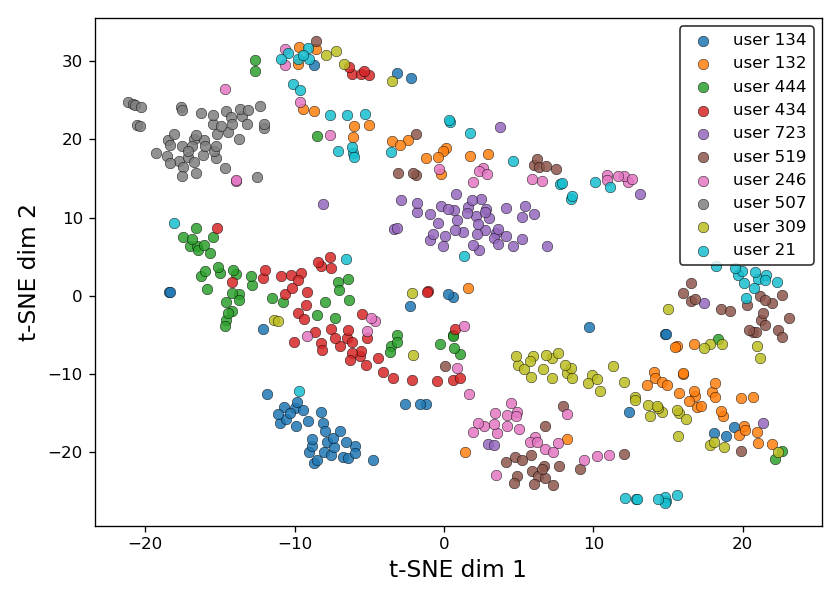}
    \caption{t-SNE of top 10 users by workout sessions.}
    \label{fig:tsne_users}
  \end{subfigure}\hfill
  \begin{subfigure}[t]{\linewidth}
    \centering
    \includegraphics[width=\linewidth]{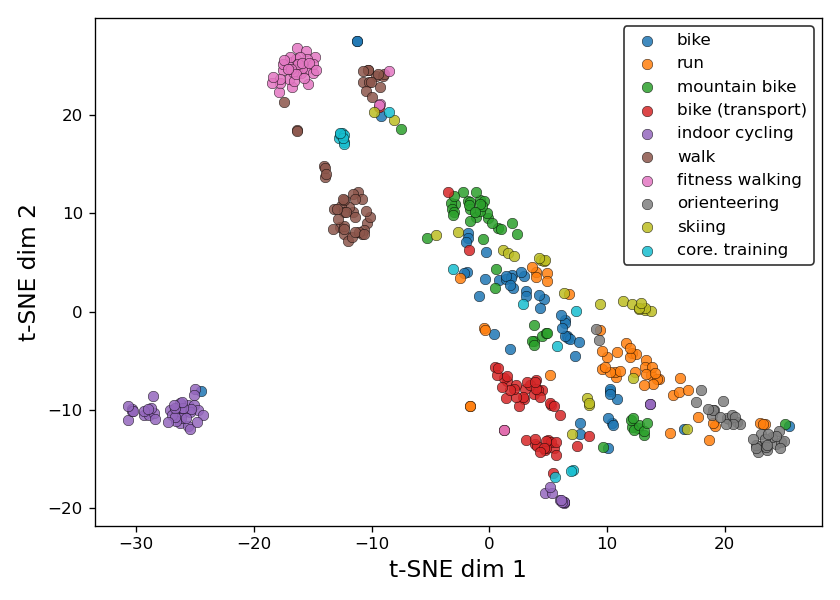}
    \caption{t-SNE of the top 10 sports.}
    \label{fig:tsne_sports}
  \end{subfigure}
   \caption{t-SNE visualizations of learned feature representations, colored by (a) user identity and (b) sport category. All plots show clear separation between different groups, indicating the effectiveness of the contrastive learning strategy in representation learning.}
  \label{fig:tsne_visualizations}
\end{figure}

\textbf{Representation Visualization.} To evaluate the effectiveness of learned representations, we first visualize user embeddings using t-SNE across different user identities and sports categories. Both Fig.~\ref{fig:tsne_users} and Fig.~\ref{fig:tsne_sports} demonstrate well-separated clusters for the top 10 users and sports categories, respectively. These visualizations confirm that the learned representations are highly discriminative across both users and sports, which is essential for personalized heart rate modeling.

\textbf{Quantitative Evaluation of Representation Discriminability.}
To quantitatively assess how well the learned representations separate different labels, we evaluate a battery of standard structure-aware metrics on both FitRec and \textsc{ParroTao}. For each representation and each label space, we measure (i) local label recovery via $1$-NN classification accuracy (kNN@1), (ii) label--representation alignment via Normalized Mutual Information (NMI) and V-measure, and (iii) cluster quality via Davies--Bouldin Index (DBI, lower is better) and a global rank-based score (RankMe). We consider three label types: \emph{sport type}, \emph{user identity}, and the combined \emph{user$\times$sport} label. The full results are reported in Table~\ref{tab:repr_merged_onecol}.

Across all label granularities and on both datasets, our model consistently achieves better kNN accuracy, stronger label–embedding alignment, and more compact, better-separated clusters than the FitRec baseline. This provides quantitative evidence that the proposed architecture learns representations that more cleanly organize sessions by both sport type and user-specific characteristics.

\begin{table}[!t]
\centering
\caption{Representation quality under different label types. Higher is better for RankMe, kNN@1, NMI, and V-measure; lower is better for DBI. Label types: S = sport, U = user, U$\times$S = user$\times$sport.}
\label{tab:repr_merged_onecol}

\scriptsize
\setlength{\tabcolsep}{3.5pt}
\renewcommand{\arraystretch}{1.02}

\resizebox{\columnwidth}{!}{%
\begin{tabular}{l l c c c c c}
\toprule
\textbf{Label} & \textbf{Model} & \textbf{RankMe} & \textbf{kNN@1} & \textbf{NMI} & \textbf{DBI$\downarrow$} & \textbf{V-meas.} \\
\midrule
\multicolumn{7}{c}{\textbf{FitRec}} \\
\midrule
S        & FitRec & 40.91 & 0.931 & 0.110 & 1.924 & 0.110 \\
S        & Ours   & \textbf{61.17} & \textbf{0.945} & \textbf{0.140} & \textbf{1.508} & \textbf{0.140} \\
U        & FitRec & 40.91 & 0.495 & 0.164 & 1.714 & 0.164 \\
U        & Ours   & \textbf{61.17} & \textbf{0.713} & \textbf{0.617} & \textbf{1.462} & \textbf{0.617} \\
U$\times$S & FitRec & 40.91 & 0.367 & 0.313 & 1.699 & 0.313 \\
U$\times$S & Ours   & \textbf{61.17} & \textbf{0.492} & \textbf{0.694} & \textbf{1.525} & \textbf{0.694} \\
\midrule
\multicolumn{7}{c}{\textbf{\textsc{ParroTao}}} \\
\midrule
S        & FitRec & 37.81 & 0.929 & 0.084 & 1.628 & 0.084 \\
S        & Ours   & \textbf{52.33} & \textbf{0.934} & \textbf{0.089} & \textbf{1.420} & \textbf{0.089} \\
U        & FitRec & 37.81 & 0.369 & 0.170 & 1.614 & 0.170 \\
U        & Ours   & \textbf{52.33} & \textbf{0.697} & \textbf{0.585} & \textbf{1.450} & \textbf{0.585} \\
U$\times$S & FitRec & 37.81 & 0.437 & 0.326 & 1.570 & 0.326 \\
U$\times$S & Ours   & \textbf{52.33} & \textbf{0.697} & \textbf{0.659} & \textbf{1.323} & \textbf{0.659} \\
\bottomrule
\end{tabular}%
}
\end{table}





\subsection{Group 4: Downstream Applications} \label{subsec:g4}

\begin{figure}[!ht]
    \centering
    \begin{subfigure}[b]{\linewidth}
        \centering
        \includegraphics[width=\linewidth]{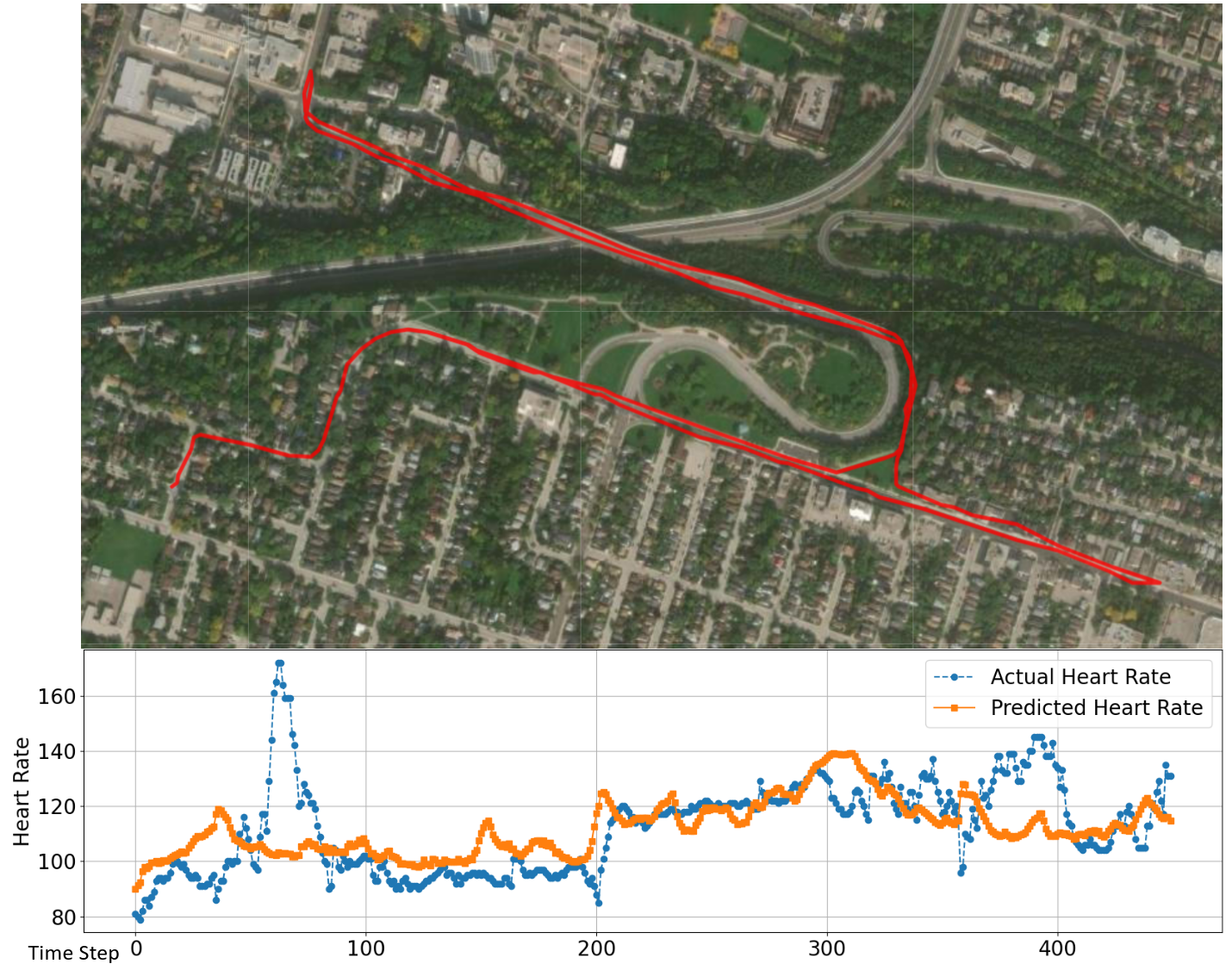}
        \caption{Candidate route A and Corresponding HR.}
        \label{fig:route1}
    \end{subfigure}
    \hfill
    \begin{subfigure}[b]{\linewidth}
        \centering
        \includegraphics[width=\linewidth]{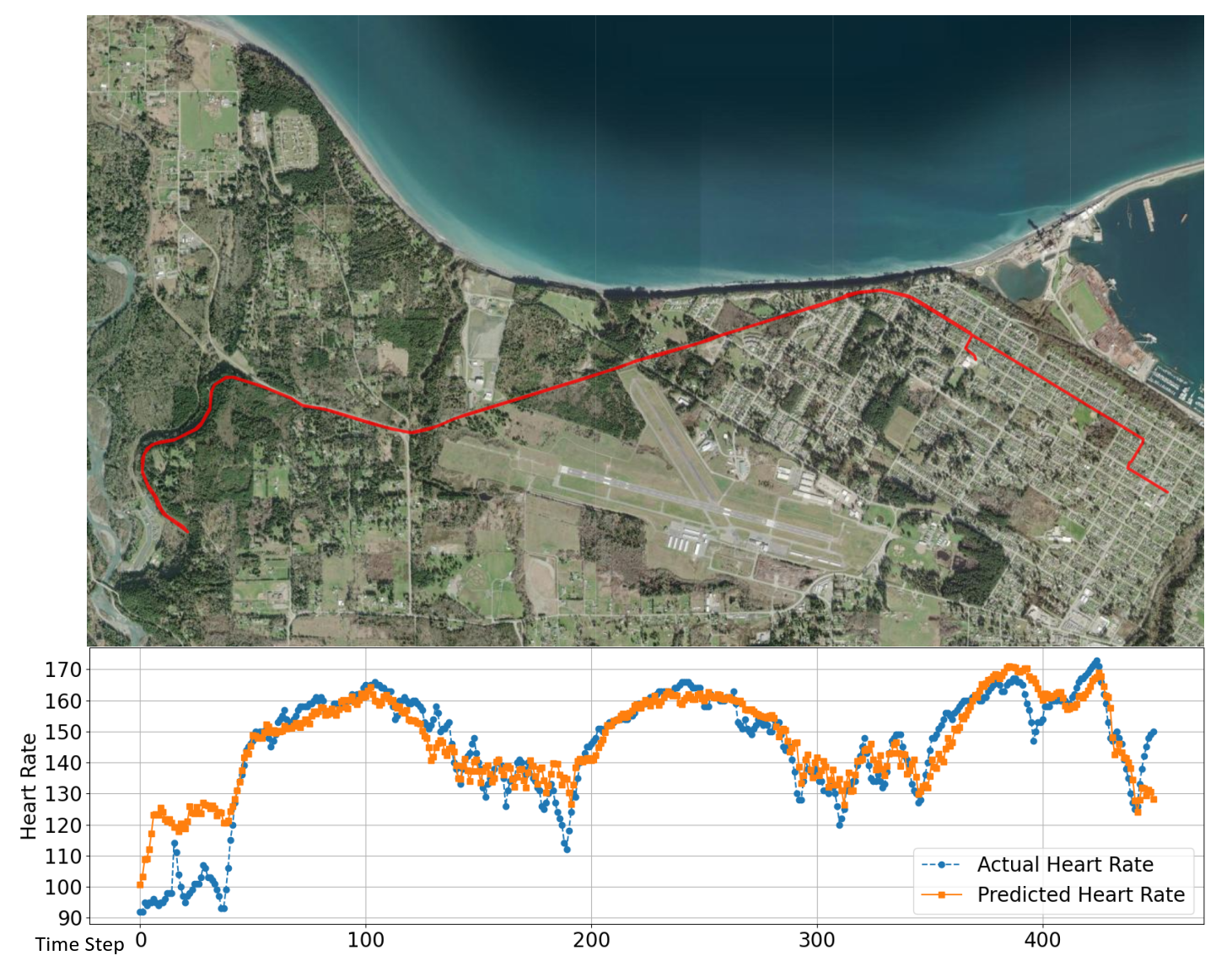}
        \caption{Candidate route B and Corresponding HR.}
        \label{figures/route2.png}
    \end{subfigure}
    \caption{Route recommendation example. (a, b) Topographical profiles and corresponding heart rate responses for two candidate routes, A and B. The close agreement validates the model’s effectiveness in forecasting physiological demands and supporting personalized route selection.}
    \label{fig:route_recommendation}
\end{figure}

Our model’s accurate heart rate prediction capabilities unlock several valuable downstream applications. Here we demonstrate two applications. 

\textbf{Personalized Route Recommendation.} Choosing a route with the appropriate difficulty is key to effective training. Our model can serve as a planning tool by forecasting the physiological demands of different routes based on topographical data and intended pace. Athletes can input their planned routes and receive full heart rate curve predictions from our model, enabling them to compare physiological costs before starting their run.

Fig.~\ref{fig:route_recommendation} shows an example of route selection in cross-country running. Two candidate routes, A and B, are considered. Route~A lies on flat terrain, whereas route~B climbs through mountainous ground.  Our model predicts that route~B induces higher average heart rate and more pronounced fluctuations compared to route~A. Subsequent measurements confirm that the actual heart rate curves closely match these predictions, demonstrating the model’s effectiveness for route recommendation. Athletes can rely on the predicted heart rate to choose the route that best matches their training goals.

\textbf{Heart Rate Imputation.}
Finally, we evaluate whether the framework can also support imputing missing heart rate (HR) values, a common issue for wrist-based sensors. We construct masked variants of the FitRec and \textsc{ParroTao} datasets 
by randomly masking short HR segments so that roughly $20\%$ of time steps are missing in each sequence. Given the original covariates, the partially masked HR sequence, and a binary mask indicating missing positions, each model is trained to reconstruct the full HR trajectory, with the loss computed only on masked time steps.

We compare our model against four baselines: (i) the FitRec model, (ii) a Kalman filter, (iii) linear interpolation, and (iv) last-observation-carried-forward (LOCF). Table~\ref{tab:hr_imputation} reports MSE on masked positions for both datasets. Our method achieves the lowest error across all settings, substantially outperforming classical interpolation and filtering methods as well as the FitRec baseline. This shows that the learned representations are not only useful for forecasting, but also for high-quality HR imputation, enabling smoother and more informative post-workout HR profiles.

\begin{table}[!t]
\centering
\caption{heart rate imputation on masked HR sequences (FitRec and \textsc{ParroTao}). We report mean $\pm$ standard deviation of MSE over masked positions (lower is better).}
\label{tab:hr_imputation}
\begin{tabular}{llc}
\toprule
Dataset & Method & MSE $\downarrow$ \\
\midrule
\multirow{5}{*}{FitRec}
 & FitRec & $69.78 \pm 0.33$ \\
 & Kalman & $832.61 \pm 1.77$ \\
 & Linear interpolation & $91.11 \pm 0.15$ \\
 & LOCF & $174.00 \pm 0.44$ \\
 & \textbf{Ours} & $\mathbf{40.84 \pm 0.17}$ \\
\midrule
\multirow{5}{*}{\textsc{ParroTao}}
 & FitRec & $19.72 \pm 0.66$ \\
 & Kalman & $167.41 \pm 0.48$ \\
 & Linear interpolation & $34.81 \pm 0.15$ \\
 & LOCF & $81.92 \pm 0.64$ \\
 & \textbf{Ours} & $\mathbf{7.54 \pm 0.08}$ \\
\bottomrule
\end{tabular}
\end{table}

\section{Conclusion} \label{sec:conclusion}


In this work, we propose a unified representation learning framework to address data heterogeneity in heart rate prediction, proposing multiple components to mitigate device dependencies and enhance representation quality. We introduce the \textsc{ParroTao} dataset to better demonstrate real-world heterogeneity challenges. Experiments on both FitRec and \textsc{ParroTao} datasets show consistent improvements across diverse devices, users, and sports scenarios, establishing a foundation for robust heart rate modeling in real-world environments.

\bibliographystyle{IEEEtran}
\bibliography{TASE/tase_reference}


\ifanonsubmission
\else
\section*{Acknowledgments}
This work was supported by the Special Funds for the Cultivation of Guangdong College Students' Scientific and Technological Innovation (Grant No. PDJH2025C12601, PDJH2026C12601). We also thank all individuals who generously provided data for this study.

\textbf{Generative AI disclosure:} We used OpenAI ChatGPT (version 5.1) for language editing across the full manuscript. The AI tool was not used to generate or modify any research content. All AI-assisted edits were reviewed and verified by the authors.
\fi
\end{document}